\documentclass[11pt,a4paper]{article}
\usepackage{emnlp2020}
\usepackage{times}
\usepackage{latexsym}

\usepackage{microtype}
\usepackage{amsmath}
\usepackage{amssymb}
\usepackage{graphicx}
\usepackage{enumitem}
\usepackage{url}
\usepackage{multirow}
\usepackage{amsthm}
\usepackage{booktabs}
\usepackage[normalem]{ulem}
\newtheoremstyle{indented}{3pt}{3pt}{\addtolength{\leftskip}{0.6em}}{}{\bfseries}{.}{.5em}{}

\theoremstyle{indented}

\title{Part \& Whole Extraction: Towards A Deep Understanding of Quantitative Facts for Percentages in Text}

\author{Lei Fang \\
  Microsoft Research Asia \\  
  \texttt{leifa@microsoft.com} \\\And
  Jian-Guang Lou \\
  Microsoft Research Asia \\  
  \texttt{jlou@microsoft.com} \\}

\date{}

\begin{document}
\maketitle

\begin{abstract}
We study the problem of quantitative facts extraction for text with percentages.
For example, given the sentence ``30 percent of Americans like watching football, while 20\% prefer to watch NBA.'', 
our goal is to obtain a deep understanding of the percentage numbers (``30 percent'' and ``20\%'') by extracting their quantitative facts: {\tt part} (\textit{``like watching football''} and \textit{``prefer to watch NBA''}) and {\tt whole} (\textit{``Americans''}).
These quantitative facts can empower new applications like automated infographic generation.
We formulate {\tt part} and {\tt whole} extraction as a sequence tagging problem. 
Due to the large gap between {\tt part}/{\tt whole} and its corresponding percentage,
we introduce skip mechanism in sequence modeling, %to address the long-term dependencies.
and achieved improved performance on both our task and the CoNLL-2003 named entity recognition task. Experimental results demonstrate that learning to skip in sequence tagging is promising.
\end{abstract}

\section{Introduction}
\label{sec.introduction}
\begin{figure}
	\centering
	\includegraphics[width=0.6\columnwidth]{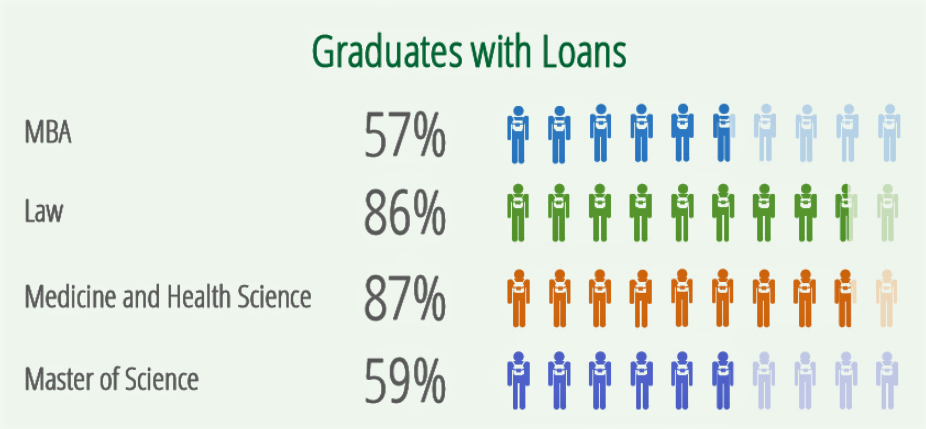}
	\caption{An Infographic Example.\cite{ex.infographic} }
	\label{fg.infographic_example}
\end{figure}

Infographics are graphic visual representations of information and data intended to present information quickly and clearly.
In the area of design and visualization, according to the survey conducted by \citet{cui2019text2viz}, about 43\% of the high-quality infographics on the web are proportion related. 
Figure~\ref{fg.infographic_example} shows an infographic example.
Infographics are more engaging and memorable than plain text, and it motivates us that how we can generate the infographics automatically from the text.% with percentage numbers.

Percentage is often used to express a proportionate \texttt{part} of a \texttt{whole}. 
Figure~\ref{fg.infographic_example} shows the percentages of graduates with loans under different degrees. 
For the four percentages, students with each degree can be considered as the corresponding {\tt wholes}, and they share the same {\tt part}  
-- ``\textit{graduates with loans}''.
Therefore, from text analytics part, to automatically generate infographics from the text, we need to extract quantitative facts
{\tt part} and {\tt whole}.
\citet{lamm2018qsrl} define the Quantitative Semantic Role Labeling (QSRL) Schema for quantitative fact extraction. 
Here, we only focus on percentage related quantitative facts {\tt part} and {\tt whole} in QSRL.
Sentences like ``\textit{The current U.S. GDP growth rate is 1.9\%.}'' are filtered out with a rule-based classifier because there is no {\tt part} or {\tt whole} associated with the percentage. 

\citet{lamm2018analogies} propose a Conditional Random Field (CRF)~\cite{lafferty2001crf} based approach to jointly perform quantitative fact extraction and relation analysis.
Following the QSRL schema defined by~\citet{lamm2018qsrl}, {\tt part} and {\tt whole} could be considered as entity-like semantic roles of the corresponding percentage.
Therefore, we formulate {\tt part} and {\tt whole} extraction as a sequence tagging task, and leverage the state-of-the-art techniques in named entity recognition (NER)~\cite{huang2015bilstmcrf,lample2016ner} or semantic role labling (SRL)~\cite{he2017srl,he2018srl}.
We regard this approach as the baseline approach and will present more details in Section~\ref{sec.baselineapp} later.

In our task, the gap between {\tt part} or {\tt whole} with its corresponding percentage could be quite large. For example, in sentence ``\textit{The World Bank estimates that 77\% of jobs in China, 69\% of jobs in India, and 85\% of jobs in Ethiopia\underline{,} are at risk of automation}~\cite{ex.wiki}.'', the three percentage numbers (77\%, 69\% and 85\%) share the same {\tt part} ``\textit{at risk of automation}'', which locates at the last sub-sentence preceded by a comma. 
The baseline approach fails to extract the {\tt part} for all the percentage numbers, mainly due to the long-term dependency. 
However, it could extract {\tt part} and {\tt whole} for all percentages correctly when the last (underlined) comma is removed.
Not all input tokens are equally important, this motivates us that we could achieve improved performance by skipping some tokens in the original text.

In this paper, we introduce skip mechanism to address the long-term dependency for {\tt part} and {\tt whole} extraction.
Our approach is generic and could be applied to other sequence tagging tasks like NER.
Experimental results show that our solution outperforms competitive strong baselines on both our task and the CoNLL-2003 NER task~\cite{conll2003}, demonstrating that learning to skip in the sequence modeling is promising.
\section{Baseline Approach}
\label{sec.baselineapp}

%We first briefly describe how we label the data before we introduce the baseline approach.
For all input text, we use {\it Recognizers-Text}\footnote{\url{https://github.com/microsoft/Recognizers-Text}}, a rule-based tool, to extract and normalize percentages. 
After that, we label the {\tt part} and {\tt whole} for each percentage.
%Figure~\ref{fg.data_labeling} shows the data labeling for the aforementioned example.
%To illustrate the {\tt part} and {\tt whole} for each percentage clearly, we  add the number \{$1$, $2$, $\cdots$\} as suffix to tags \{``{\em Percentage}'', ``{\em Part}'', ``{\em Whole}'' \}.
%\begin{figure}[!htbp] 
%	\centering
%	%\includegraphics[width=0.9\linewidth]{datalabeling.png}
%	 \includegraphics[clip, trim=6cm 9.8cm 6cm 8cm, width=0.5\textwidth]{data_labeling.pdf}
%	 % trim={<left> <lower> <right> <upper>}
%	 % use \fbox{ } to tune the size
%	\caption{Data Labeling Example}
%	\label{fg.data_labeling}
%\end{figure}
%
Given the percentage, {\tt part} and {\tt whole} could be considered as entity-like semantic roles. 
Following~\citet{lample2016ner} in NER, we design a bidirectional Long Short Term Memory (biLSTM)~\cite{Sepp1997LSTM} with CRF as the extraction model, {\tt part} and {\tt whole} are text spans represented using ``BIOUL'' (same as ``BIOES'') schema\footnote{We will use this schema through this paper as previous studies~\cite{ratinov2009design,Enhancing2015} have reported meaningful improvement with this scheme in NER. }.
We encode the words in a sentence with fixed pre-trained word embeddings.
We also concatenate some additional features such as Part-of-Speech (POS) tags and whether a token is the percentage token (based on the output of {\it Recognizers-Text}).
These features are randomly-initialized, learned embeddings.

In semantic role labeling, for sentences with multiple predicates, \citet{he2017srl} concatenate the input features with a binary mask \{$0$, $1$\} indicating whether the word is of the current predicate.
After that, sentences are fed to the model multiple times, each time with a mask $1$ for one predicate.
Similar to~\citet{he2017srl}, we append the input feature with a binary mask indicating whether it is the token of the current percentage for extraction, and feed the model multiple times to obtain the {\tt parts} and {\tt wholes} for all percentages.
We regard this approach as the baseline approach. It should be noted that the backbone of the baseline approach is the biLSTM+CRF model, for simplicity, we denote the baseline approach as biLSTM+CRF.

\section{Learning to Skip}
\subsection{Overview}
biLSTM+CRF is one of the state-of-the-art approaches in sequence tagging,
where LSTM is a special kind of recurrent neural network (RNN) that designed to capture long-range dependencies. 
LSTM has become a popular architecture for modeling natural language.
It reads every input token and outputs a distributed representation for each token.
However, not all input tokens are equally important in many NLP tasks, and it is the fact that texts are often written redundantly.
%For example, the attention mechanism~\cite{BahdanauCB14Attention} that addresses the importance of each token, has achieved remarkable success in various NLP tasks.
For this reason, many variants of RNNs, including LSTM-Jump~\cite{yu2017learning}, Skim-RNN~\cite{seo2018neural}, Skip RNN~\cite{campos2018skip}, Structural-Jump-LSTM~\cite{hansen2018neural}, and Leap-LSTM~\cite{Leap-LSTM}, are proposed to skip/skim some uninformative tokens.

\begin{figure} 
	\centering
	 \includegraphics[clip, trim=6cm 5.5cm 6cm 5.8cm, width=0.8\columnwidth]{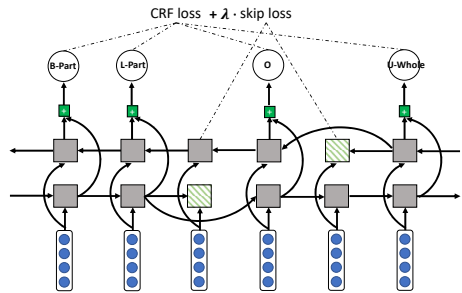}
	 % trim={<left> <lower> <right> <upper>}
	 % use \fbox{ } to tune the size
	\caption{Skip model overview, the slash squares represent the skipped tokens.}
	\label{fg.model_overview}
\end{figure}

A natural way to empower the biLSTM+CRF model with the capability to skip is to replace the LSTM layer with its skip variants.
Figure~\ref{fg.model_overview} presents an overview of our proposed model.
Since there are two LSTM layers (forward and backward), a token is skipped if either the forward layer or the backward layer decides to skip it.
The output of remained tokens is fed into the CRF layer.
The model is jointly optimized based on the CRF loss on the remained tokens and the skip loss on the skipped tokens.
In this paper, we adopt the Skip LSTM proposed by ~\citet{campos2018skip} which has two advantages over other skip variants, 1) Skip LSTM only augments the LSTM with a linear layer, the size of the additional parameter is rather small; 2) training Skip LSTM is fairly simple, no reinforcement learning techniques are needed. 

\subsection{Skip LSTM}
Here, we briefly introduce the Skip LSTM, please refer to \cite{campos2018skip} for details.
Let $\mathbf{x}=(x_1,\cdots,x_{T})$ denote the input sequence.
LSTM generates a state sequence $\mathbf{s}=(s_1,\cdots,s_{T})$ by applying the state transition model $S$:% from $t=1$ to $T$:
\begin{equation}
s_t = S(s_{t-1}, x_t)
\end{equation}

The Skip LSTM augments LSTM with a binary state update gate, $u_t \in \{0,1\}$, 
where $u_t=1$ means that the state will be updated using the state transition model $S$ (as the same in LSTM), 
$u_t=0$ denotes that the state is copied from the previous time step, and the input token at timestep $t$ is skipped.
%In Skip LSTM, 
The binary state update gate $u_t$ is calculated by
\begin{displaymath}
\begin{split}
&u_t = f_{binarize}(\tilde{u}_t)\\
&s_t = u_t \cdot S(s_{t-1}, x_t) + (1-u_t) \cdot s_{t-1}\\
&\Delta \tilde{u}_t = \sigma(W_ps_t+b_p) \\
&\tilde{u}_{t+1} = u_t\cdot\Delta\tilde{u}_t + (1-u_t)\cdot min(\tilde{u}_t+\Delta\tilde{u}_t,1)
\end{split}
\end{displaymath}
where $W_p$ is a weights vector, $b_p$ is a scalar bias, 
$\sigma$ is the sigmoiod function and $f_{binarize}:[0,1]\rightarrow\{0,1\}$ binarizes the input value.
In Skip LSTM, $f_{binarize}$ is implemented as a determinstic function $u_t = round(\tilde{u}_t)$.
The model is differentiable except $f_{binarize}$, following \cite{campos2018skip}, we define the gradients during backward pass as:
\begin{equation}
\frac{ \partial f_{binarize}(x)}{\partial x}=1
\end{equation}
This is a straight-through estimator~\cite{straight-estimator}. Though it is a biased estimator, it is more efficient than other unbiased but high-variance estimators such as REINFORCE~\cite{Williams1992}.
Skip LSTM copies the previous states of the skipped tokens, which could increase its ability to handle long-term dependencies.

\subsection{Bidirectional Skip LSTM + CRF}
Following the biLSTM+CRF, we stack two layers of Skip LSTMs (forward and backward) with CRF for sequence tagging. 
A token is skipped if either the forward or backward Skip LSTM decides to skip it. 
The Skip LSTM outputs of the remained tokens are fed into the CRF layer.
The CRF loss (the negative log-probability of the correct tag sequence), $\mathcal{L}_{\text{CRF}}$, is calculated based on the remained tokens. 
In order to prevent our model from skipping entity tokens, we introduce skip loss on the skipped tokens. Let $\mathbf{y}=(y_1,\cdots,y_{T})$ denote tag sequence of $\mathbf{x}$,  the skip loss on $\mathbf{x}$ is calculated by 
\begin{equation}
\mathcal{L}_{\text{skip}}(\mathbf{x},\mathbf{y})=\sum\nolimits_{t}\sum\nolimits_{d}(1-u_{t,d}) \cdot \textbf{ I}_{\Omega}(y_t)
\end{equation}
where $d$ is the direction, $d \in \{\text{forward}, \text{backward}\}$, $u_{t,d}$ is the update gate of the token $x_t$ on direction $d$, $\Omega$ is the set of entity related tags, \textit{e}.\textit{g}., $\Omega$ contains all the tags except 'O' in ``BIOUL'' tagging schema, \textbf{ I} is the indicator function, it equals to $1$ when $y_t \in \Omega$ and $0$ otherwise. The training objective is to jointly optimize the CRF loss and skip loss on the training set $\mathcal{D}$, which can be formulated as
\begin{equation}
\mathcal{L}_{\mathcal{D}}=\mathbb{E}_{(\mathbf{x},\mathbf{y})\thicksim \mathcal{D}}[\mathcal{L}_{\text{CRF}}(\mathbf{x},\mathbf{y})+\lambda \mathcal{L}_{\text{skip}}(\mathbf{x},\mathbf{y})]
\end{equation}
$\lambda$ balances the weight between CRF loss and skip loss.
For prediction, tags of the skipped tokens will be 'O', except those tokens that exist in the predicted entity span. For example, an entity has three tokens, and our model skips the second one, CRF assigns 'B-Entity' and 'L-Entity' to the first and third token. We will set the tag of the second token as 'I-Entity' to obtain valid entity spans.
%\section{Experiments\footnote{Code will be publicly available, reviewers could reproduce the results with the submitted codes and models.}}
\label{sec.exp}
\subsection{Data \& Settings}
\label{sec.exp.setting}
For {\tt part} and {\tt whole} extraction, we sample and label $1,423$ sentences with percentages from Wikipedia, where the percentages are extracted and normalized using {\it Recognizers-Text}.
On average, there are $2.23$ percentages per sentence.
The train/dev/test split of sentences (instances) is $825$($1,798$)/$278$($598$)/$282$($592$), note that sentence with multiple percentages will be converted to multiple instances as explained in Section~\ref{sec.baselineapp}. 

For both baseline approach (biLSTM+CRF) and our model, we use GloVe~\cite{pennington2014glove} as the pre-trained word embeddings, the length of the POS tag feature is $25$,
the dimension of the hidden state is $50$, and the learning rate is $0.001$.

To verify the effectiveness of skipping in sequence modeling, we run our model on the CoNLL-2003 NER task. 
%This dataset contains four different types of entities: locations, persons, organizations, and miscellaneous entities.
We follow the AllenNLP~\cite{gardner2018allennlp} settings\footnote{\url{https://github.com/allenai/allennlp/tree/v0.8.3/training\_config}}
of using GloVe (denoted by Allen-GloVe) and ELMo~\cite{elmo2018} (denoted by Allen-ELMo) on the NER task with biLSTM-CNNs-CRF~\cite{chiu2016,ma2016,peters2017semi}, where CNN is employed to model character-level information. 
We set the number of biLSTM (bi-Skip LSTM) in Allen-GloVe and Allen-ELMo to one\footnote{There are two biLSTM layers in \cite{elmo2018}.}.
It should be noted that experiments on NER is to demonstrate that our model could also be applied to other sequence tagging tasks. 
Boosting the performance of NER is not the focus of our work.

In our model, $\lambda$ balances the importance between CRF loss and skip loss. We select $\lambda$ ranging from $0.02$ to $1.00$, with step size $0.02$. 
For the baseline and our model (under each $\lambda$), the results are averaged over 20 runs with random initialization.

\subsection{{\tt part} and {\tt whole} Extraction}
\begin{table}
\centering
\resizebox{0.9\columnwidth}{!}{
\begin{tabular}{cclll}  
\toprule
			&    			  & overall		& {\tt part}   & {\tt whole} \\
\midrule
\multicolumn{2}{c}{biLSTM+CRF}& 72.13(0.61) &  70.47(0.88) & 73.83(1.01)\\ \hline
\multirow{2}{*}{+skip}&best    & 73.75(0.75) &  72.17(0.84) & 75.38(1.04)\\
					 &median  & 73.38(0.55) &  72.05(0.83) & 74.72(0.89)\\
%					 &worst   & 72.99(1.02) &  71.61(1.24) & 74.41(1.17)\\
%					 &mean    & 73.38       &  71.97       &  74.81     \\
\bottomrule
\end{tabular}}
\caption{F1 of {\tt part} \& {\tt whole} Extraction.}
\label{tab:prex}
\end{table}
Our approach outperforms the biLSTM+CRF significantly  ($p$-value$<\!\!0.01$) under all settings of $\lambda$.
Table~\ref{tab:prex} shows the results with standard deviation.
``best'' (``median'') is the best (median) of the averaged results when varying the parameter $\lambda$. It should be noted that no existing approaches could be applied to our task, and biLSTM+CRF is a strong baseline. 
For each $\lambda$, the result is calcuated over 20 runs.
%``mean'' in Table~\ref{tab:prex} is the mean of averaged F1 scores over all $\lambda$.
The results indicate that learning to skip in sequence modeling improves the {\tt part} and {\tt whole} extraction performance.

We find that our model is stable when varying $\lambda$, because the number of skipped tokens is rather small. On average, we skipped $24.6$ tokens on the test set, among which $1.53$ are entity tokens (skip errors), while the total number of tokens is $17,749$.
We do further analysis by training the standard biLSTM+CRF with some stopwords and punctuations randomly removed.
The performance drops significantly, which explains that the skip mechanism should not be activated frequently.

To obtain an overview of the skipped tokens, we aggregate the skipped tokens on the test set over all the settings, and the most commonly skipped tokens are 
%\{``,'',``.'',``to'',``of'',``and'',``in'',``by'',``the'',``were'',``-''\}
\{, . to of and in by the were -\}\footnote{We sort the tokens based on the skip frequency divided by the logarithm of frequency on the test set.}, which demonstrates that our approach could effectively skip some uninformative tokens in text sequence.

%We review the extraction output of our approach on test data, 
%and find that some of the results are acceptable even when the predicted spans does not match the labeled ones. 
Our approach fails when the multiple percentages in the sentence are expressed with ``respectively''.
%For example,  ``South Asians include Indian, Pakistani, and Nepalese, who respectively made up 0.4\%, 0.3\%, and 0.2\% of Hong Kong's population in 2011''. 
For example,  \textit{Indian, Pakistani, and Nepalese, made up 0.4\%, 0.3\%, and 0.2\% of Hong Kong's population, respectively}. 
There are also some other complicated cases (ellipsis, coreference, etc.) that both our approach and baselines fails, which cannot be summarized here due to space limit. 
We leave them as our future work.

\subsection{CoNLL-2003 NER task} 
\begin{table}
\centering
\resizebox{0.65\columnwidth}{!}{
\begin{tabular}{c|l|l}  
\toprule 
\multirow{3}{*}{Allen-GloVe}& \multicolumn{2}{r}{89.80(0.27)} \\ \cline{2-3}
					 &+skip(best)    & \textbf{89.95(0.21)} \\
					 &+skip(median)  & \textbf{89.86(0.25)} \\ \hline
%					 &worst   & 89.71(0.38) \\
%					 &averaged& 89.84       \\ \hline
\multirow{3}{*}{Allen-ELMo}& \multicolumn{2}{r}{92.26(0.16)} \\ \cline{2-3}
					 &+skip(best)    & \textbf{92.43(0.14)} \\
					 &+skip(median)  & \textbf{92.34(0.18)} \\ \hline
%					 &worst   & 92.29(0.20) \\
%					 &averaged& 92.34       \\ \hline
%\multicolumn{2}{c|}{\cite{lstm-dskip2019}}& 91.56(-)       \\ \hline
%\multicolumn{2}{c|}{\cite{tran-etal-2017-named}}& 91.66(-)       \\ \hline
\multicolumn{2}{c|}{\cite{peters2017semi}}& 91.93(0.19)       \\ \hline
\multicolumn{2}{c|}{\cite{elmo2018}}& 92.22(0.10)       \\ 
\bottomrule
\end{tabular}}
\caption{F1 score on CoNLL-2003 NER task.}
\label{tab:ner}
\end{table}
On the CoNLL-2003 NER task, for Allen-GloVe, when introducing skipping, $44$ out of $50$ settings of $\lambda$ achieve improvements over the biLSTM-CNNs-CRF, and for Allen-ELMo, our approach outperforms baseline under all settings of $\lambda$.   
Table~\ref{tab:ner} shows the results with several strong baselines using the same version of GloVe. We do not compare with BERT~\cite{bert} or flair~\cite{flair2018} because we focus on skipping in biLSTM+CRF.
%The results also demonstrate that learning to skip is promising in sequence modeling.
The improvement is not as significant as in {\tt part} and {\tt whole} extraction, the reason might be that dependencies among entities are weaker than that among {\tt part} and {\tt whole}.

The number of skipped tokens is also quite small on the test set. When introducing skipping to Allen-ELMo, our model skips 16.7 tokens on average,
among which $0.57$ are entity tokens, while the total number of tokens is $46,435$.
%8.64 1.36
%We collect the skipped tokens in a similar way as in {\tt part} and {\tt whole} extraction.
We find that those commonly skipped tokens are mostly stopwords, for example, in Allen-ELMo with skip, the most common skipped tokens are \{- , said 's was of has and the in\}, which demonstrates that our approach could skip uninformative tokens effectively. 
\section{Conclusion }
We study {\tt part} and {\tt whole} extraction for percentages in text. 
The output of our work can support new applications like automated graphic plot generation.
Experimental results on both our task and NER show that learning to skip is effective and promising in sequence tagging tasks.
%In the future, %we plan to investigate other skip mechanisms, 
%and in addition to {\tt part} and {\tt whole}, 
%we plan to perform other quantitative facts extraction to support more types of automated graphic plot generation. 

\bibliography{emnlp2020}
\bibliographystyle{acl_natbib}

\end{document}